# Evolutionary Algorithms[*]

**David W. Corne and Michael A. Lones**

School of Mathematical and Computer Sciences

Heriot-Watt University, Edinburgh, EH14 4AS

## 1. Introduction

Evolutionary algorithms (EAs) are population-based metaheuristics. Historically, the design of EAs was motivated by observations about natural evolution in biological populations. Recent varieties of EA tend to include a broad mixture of influences in their design, although biological terminology is still in common use. The term 'EA' is also sometimes extended to algorithms that are motivated by population-based aspects of EAs, but which are not directly descended from traditional EAs, such as scatter search. The term evolutionary computation is also used to refer to EAs, but usually as a generic term that includes optimisation algorithms motivated by other natural processes, such as particle swarm optimisation and artificial immune systems. Although these algorithms often resemble EAs, this is not always the case, and they will not generally be discussed in this chapter. For a discussion of their commonalities and differences, the reader is referred to [1].

Over the years, EAs have become an extremely rich and diverse field of study, and the sheer number of publications in this area can create challenges to people new to the field. To address this, this chapter aims to give a concise overview of EAs and their application, with an emphasis on contemporary rather than historical usage.

The main classes of EA in contemporary usage are (in order of popularity) genetic algorithms (GAs), evolution strategies (ESs), differential evolution (DE) and estimation of distribution algorithms (EDAs). Multi-objective evolutionary algorithms (MOEAs), which generalise EAs to the multiple objective case, and memetic algorithms (MAs), which hybridise EAs with local search, are also popular, particularly within applied work. Special-purpose EAs, such as genetic programming (GP) and learning classifier systems (LCS) are also widely used. These are all discussed in this chapter.

Although these algorithms differ from each other in a number of respects, they are all based around the same core process. Each of them maintains a population of search points (known variously as

---

candidate solutions, individuals, chromosomes or agents). These are typically generated at random, and are then iteratively evolved over a series of generations by applying variation operators and selection. Variation operators generate changes to members of the population, i.e. they carry out moves through the search space. After each generation, the objective value (or *fitness*) of each search point is calculated. Selection then removes the search points with the lowest objective values, meaning that only the best search points are maintained, and new search points are always derived from these. It is this combination of maintaining a population of search points and carrying out selection between search points that differentiates EAs from most other metaheuristics.

Each EA uses its own distinctive set of variation operators, which are sometimes inspired by the mutative and recombinative processes that generate diversity in biological evolution. The mutation operator resembles the generation of 'moves' in other optimisation algorithms, and involves sampling the neighbourhood around an existing search point in some fashion. A typical approach would be to randomly change one component of a solution, though a particular EA may use more than one kind of mutation operator. The recombination (or *crossover*) operator explores the region between two or more search points, for example by randomly reassembling the components that make up two existing solutions. This process of searching the region between existing search points is also a distinctive feature of EAs, though its practical utility depends upon the structure of the search space. Some EAs, particularly 'evolutionary programming' and older varieties of evolution strategy, do not use recombination at all.

## 2. Principal Algorithms

In this section, we provide brief introductions to the principal classes of EA that are in current use, and then discuss existing understanding of their performance and applicability.

### 2.1 Genetic Algorithms

Genetic algorithms, or GAs, are one of the earliest forms of EA, and remain widely used. Candidate solutions, often referred to as *chromosomes* in the GA literature, comprise a vector of decision variables. Nowadays, these variables tend to have a direct mapping to an optimisation domain, with each decision variable (or *gene*) in the GA chromosome representing a value (or *allele*) that is to be optimised. However, it should be noted that historically GAs worked with binary strings, with real values encoded by multiple binary symbols, and that this practice is still sometimes used. GA solution vectors are either fixed-length or variable-length, with the former the more common of the two.

Given their long history, genetic algorithm implementations vary considerably. However, it is fairly common to use a mutation operator that changes each decision variable with a certain probability (values of 4-8% are typical, depending upon the problem domain). When the solution vector is a binary string, the effect of the mutation operator is simply to flip the value. More generally, if the solution vector is a '$k$-ary' string, in which each position can take any of a discrete set of $k$ possible values, then the mutation operator is usually designed to choose a random new value from the available alphabet. If the solution vector is a string of real-valued parameters within a set range, the new value may be sampled from a uniform distribution in that range, or it may be sampled from a non-uniform (e.g. Gaussian) probability distribution centred around the current value. The latter is generally the preferred approach, since it leads to less disruptive change on average. Recombination is typically implemented using two-point or uniform crossover. Two-point crossover chooses two *parent* solutions and two *crossover points* within the solutions. The values of the decision variables lying between these two points are then swapped to form two *child* solutions. Uniform crossover is similar, except that crossover points are created at each decision variable with a given probability. Other forms of crossover have also been used in GAs. Examples include line crossover and multi-parent crossover. Other variation operators, such as inversion, have been found useful for some problems.

Various forms of selection are used with GAs. Rank-based or tournament selection are generally preferred, since they maintain exploration better than the more traditional fitness-proportionate selection (e.g. roulette-wheel selection). Note, however, that the latter is still widely used. Rank-based selection involves ranking the population in terms of objective value. Population members are then chosen to become parents with a probability proportional to their rank. In tournament selection, a small group of solutions (typically 3 or 4) are uniformly sampled from the population, and those with the highest objective value(s) become the parent(s) of the next child solution that is created. Tournament selection allows selective pressure to be easily varied by adjusting the tournament size.

## 2.2 Evolution Strategies

Evolution strategies, or ESs, also have a long history, and this parallels the development of GAs. Whilst early ESs were restricted to a single search point and used no recombination operator, modern formulations have converged towards the GA norms, and tend to use both a population of search points and recombination. A lasting difference, however, is how they carry out mutation, with ESs using *strategies* that guide how the mutation operator is applied to each decision variable. Unlike GAs, ESs mutate every decision variable at each application of the operator, and do so according to a set of *strategy parameters* that determine the magnitude of these changes. Strategy parameters usually control characteristics of probability distributions from which the new values of decision variables are drawn.

It is standard practice to adapt strategy parameters over the course of an ES run, the basic idea being that different types of move will be beneficial at different stages of search. Various techniques have been used to achieve this adaptation. Some of these involve applying a simple formula, e.g. the 1/5$^{th}$ rule, which involves increasing or decreasing the magnitude of changes based on the number of successful mutations that have recently been observed. Others are based around the idea of self-adaptation, which involves encoding the strategy parameters as additional decision variables, and hence allowing evolution to come up with appropriate values. However, the most widely used contemporary approach is covariance matrix adaptation (CMA-ES), which uses a mechanism for estimating the directions of productive gradients within the search space, and then applying moves in those directions. In this respect, CMA-ES has similarities with gradient-based optimisation methods.

ESs use different recombination operators to GAs, and often use more than two parents to create each child solution. For example, intermediate recombination gives a child solution the average values of each decision variable in each of the parent solutions. Weighted multi-recombination is similar, but uses a weighted average, based on the fitness of each parent. Also unlike GAs, ESs tend to use deterministic rather than probabilistic selection mechanisms, whereby the best solutions in the population are always used as parents of the next generation.

**2.3 Estimation of Distribution Algorithms**

Like ESs, estimation of distribution algorithms [2], or EDAs, make use of probability distributions. However, rather than using them to describe a distribution of next moves, as an ES does, EDAs use them to describe a distribution of next sample solutions. The basic mechanism is quite simple. As for most EAs, the initial population of solutions is (typically) sampled from a uniform distribution. Selection is then used to remove the poorer members of this population, and a probability distribution is then constructed that attempts to model the statistics of the relatively high-fitness sample solutions that remain in the population. Importantly, this distribution is constructed in such a way that it 'generalises' the population members suitably well. The next generation of solutions is then constructed by sampling from this distribution. So, if the distribution was highly peaked around the existing samples, for example, the next generation would explore very little beyond the previous one. This pattern of building a distribution, sampling, and selection is then iterated in the usual generational fashion, with the hope that the final probability distribution will characterise solutions that are, or are close to, globally optimal.

Whilst an EDA may be used with any kind of probability distribution, in practice it is necessary to choose a distribution that induces an appropriate trade-off between efficiency and expressiveness. More expressive models, such as Bayesian networks and Markov models, can capture dependencies between decision variables, but can be expensive to construct and sample from. Simple univariate distributions, by comparison, are cheap to build and sample from, but are unable to capture dependencies between variables. This trade-off is reflected in the range of EDAs in common use.

Population-based incremental learning (PBIL) and the compact genetic algorithm (CGA) are both examples of computationally efficient EDAs that build simple univariate models based on discrete variables. Because of their simplicity, they can be applied to large problem instances. Bayesian optimization algorithms (BOAs) lie at the other end of the spectrum: these can express dependencies between variables, and certain varieties can be applied to both discrete and continuous variables, but they are far more demanding of computational resources.

**2.4 Differential Evolution**

Differential evolution [3, 4], or DE, is a relatively recent EA formulation which uses a mechanism for adaptive search that does not make use of probability distributions. Whilst its basic mechanism is similar to a GA, its mutation operator is quite different, using a geometric approach that is motivated by the moves performed in the Nelder Mead simplex search method. This involves selecting two existing search points from the population, taking their vector difference, scaling this by a constant $F$, and then adding this to a third search point, again sampled randomly from the population. Following mutation, DE's crossover operator recombines the mutated search point (the *mutant vector*) with another existing search point (the *target vector*), replacing it if the child solution (known as a *trial vector*) is of equal or greater objective value. There are two standard forms of crossover [5]: exponential crossover and binomial crossover, which closely resemble GA two-point crossover and uniform crossover, respectively. The comparisons between target vector and trial vector play the same role as the selection mechanism in a GA or ES. Since DE requires each existing solution to be used once as a target vector, the whole population is replaced in the course of applying crossover.

An advantage of using simplex-like mutations in DE is that the algorithm is largely self-adapting, with moves automatically becoming smaller in each dimension as the population converges. More generally, the authors of the method have claimed that this sort of self-adaptation means that the size and direction of moves are automatically matched to the search landscape, a phenomenon they term *contour matching*. When compared to CMA-ES, for example, this means that the algorithm has few parameters and is relatively easy to implement.

**2.5 Performance Comparisons**

Fair comparisons of optimisation algorithms are inherently challenging [6], and arguably unachievable. Nevertheless, there have been some attempts to understand the comparative performance of different EAs, particularly within the domain of continuous optimisation. In particular, a series of workshops held at two of the largest annual EA conferences, CEC and GECCO, have sought to define benchmark suites of real-valued function optimisation problems suitable for comparing EAs (and other optimisers) [7-9]. Using these benchmarks, a number of authors have shown their algorithms to perform better than others, including variants of CMA-ES and DE (see [3]).

It should be borne in mind that these are not exhaustive studies, either in terms of problems or approaches. The 'No Free Lunch theorem' (NFLT) [10] may also be considered when attempting to generalise these results to a wider spectrum of problems, although in itself, the NFLT does not apply in the case of comparisons based on the standard suites of test problems (since those suites are not closed under permutation [11]). A nice example of the perils of comparison study in this field is shown by a recent study that showed how quite different conclusions could be drawn from a comparative study by changing minimisation problems into maximisation problems [12].

## 3. Common Variants

The general purpose EAs introduced in the last section are applicable to a wide range of problems. However, over the course of EA history, algorithmic variants have been developed to deal with the characteristics of particular categories of problem. Some of these categories are quite broad, for example problems with multiple solutions. Others are more specific, such as discrete optimisation problems. In this section, we discuss a number of these EA variants, focussing on those which are commonly used to solve real world optimisation problems.

### 3.1 Alternative Representations

In common with other optimisation algorithms, most EAs are designed to work with and optimise vectors (or, equivalently, lists or arrays) of decision variables. Solutions to many kinds of problems can be represented, either directly or indirectly, in this form. However, EAs are not limited to working with vectors, and there are often advantages to working directly with representations that are more natural for the problem domain: for example, matrices [13], trees [14], graphs [15], rule sets, etc. The general approach is the same as for the EAs discussed in the previous section, except that specialised initialisation routines and variation operators are used to randomly create, mutate and recombine instances of the appropriate solution representation. These variants are typically based around GAs or ESs, since these two classes of EA can be readily adapted to use alternative solution representations. Nevertheless, DE [16] and EDA [17] have also been used successfully with other representations.

Genetic programming (GP) [14] is a well-known GA variant that uses trees as its solution representation. GP is mostly used to optimise computer programs or mathematical expressions, but can be used for any problem that requires tree-structured solutions. A common use of GP is *symbolic regression*, which involves finding a mathematical expression that fits a particular data set. Unlike standard mathematical approaches to regression, such as curve fitting, GP makes relatively few assumptions about the function that generated the data, allowing a wide exploration of the space of possible solutions. GP is also widely used for solving classification problems. More generally, the GP community is interested in automatic programming, i.e. finding computer programs that solve a particular task, and there are many variants of GP that use particular forms of program representation. See [18] and [19] for overviews.

Some EAs work with two different solution representations, using one of these when creating and manipulating search points, the other when evaluating search points, and a mapping process that converts the former into the latter [20]. Many of these approaches are motivated by biology, and hence this process is known as a *genotype-phenotype* mapping, with the representation used during search termed the *genotype*, and the representation used for evaluation termed the *phenotype*. This approach can be used when the natural representation for a domain is not well suited to being evolved, i.e. where mutation and recombination do not lead to productive solutions. This approach has also been widely used for generating complex structures, such as large neural networks [21], where a genotype representation can be chosen that compresses repetitive features such as symmetry and modularity. This is arguably an area in which EAs benefit from their relationship to biology, since biology provides a ready source of information on how to represent complex structures in an evolvable way.

### 3.2 Hybridisation with Local Search

EAs are often considered to be global search algorithms, since they explore a relatively wide region of the search space and are relatively good at escaping local optima. However, their convergence to optimal solutions can be relatively slow when compared to local search algorithms. For this reason, EAs are often hybridised with local search, using it to locally optimise members of the population at regular intervals, hence speeding up convergence. Although the resulting hybrid algorithms are known by various names, the term *memetic algorithm* [22] (MA) has become popular in recent years. Memetic, in this case, refers to an analogy between the role of local search in these algorithms and the role of within-generation learning in biological systems, though the majority of memetic algorithms have no particular biological justification beyond this. In principle, these algorithms may involve hybridising an EA with any or with multiple local search algorithms, and consequently are very diverse. For a recent review, see [23].

Beyond hastening convergence, MAs are also seen as a means of introducing domain knowledge into EAs. This is done through the use of specialised local search operators that are relevant to a particular domain. For example, this approach underlies the success of MAs in the area of discrete optimisation [24]. Related to the idea of problem specialisation in memetic algorithms is the concept of hyperheuristics in EAs, which has developed some traction in recent years [25]. Generally speaking, hyperheuristics are applicable to domains in which a variety of so-called 'low level' heuristics exist (or can be invented) to build quick, good solutions. In the job-shop scheduling problem, for example, 'shortest-process-time' and 'earliest-available-machine' are two examples of low-level 'dispatch' heuristics that, when iterated, can build a single solution quickly. Hyper-heuristics are essentially mechanisms used to explore *combinations* of such lower-level heuristic strategies. The term hyper-heuristics is also used to describe the cases in which an EA (often GP) both creates anew, and combines, such low-level heuristics. In a nutshell, the broad idea of hyperheuristics is to search a space of algorithms that can solve a class of problems, rather than search the space of solutions directly for a single problem instance.

### 3.3 Multimodal Optimisation

An advantage of maintaining a population of search points is that EAs can be readily applied to multimodal optimisation problems in which there is more than one solution of interest. However, effective multimodal optimisation generally requires some modification to the EA's underlying behaviour, since although EAs explore diverse areas of the search space, they eventually converge to fairly small areas. This behaviour can be mitigated, to an extent, by varying the global selection pressure used when choosing parents; for example, in the case of tournament selection, the tournament size can be made small, increasing the likelihood that less fit members of the population will contribute to the next generation of search points. Whilst this increases exploration, it decreases exploitation: meaning that multiple solutions may be found, but they are less likely to be optimal. Since EAs are stochastic, and there is the potential for them to converge on different optima during different runs, another simple approach to finding multiple solutions is to run an EA multiple times. However, there is no guarantee that all optima will be explored, and algorithmic biases (such as the manner in which the initial population is generated) may favour some solutions over others.

A more effective approach is to use some kind of *niching* technique [26]. These aim to preserve global diversity in the population, but without lowering local selective pressure. Niching approaches are motivated by the biological concept of evolutionary niches, in which species compete within a niche but not between niches. In optimisation terms, a niche is a local region within the search space that contains a solution of interest, and the aim is for the population to be distributed across all the relevant niches. Niching has been studied for some time in GAs, and techniques include crowding [27], fitness sharing [28], spatial segregation [29] and clustering [13]. For comparative studies, see [5] and [7]. A simple but effective example of niching is probabilistic crowding [27]. This works at the operator level, and always replaces parent solutions with their children, meaning that search points are usually replaced with nearby search points and the population remains spread across the search space. Similar techniques have also been developed for use in DE. Niching is less commonly used in ESs, in part due to their use of smaller populations, though examples do exist [30]. It is also common to use multi-objective evolutionary algorithms (see below) to solve multimodal problems, since these algorithms often have effective mechanisms for preserving population diversity.

## 3.4 Multi-objective Optimisation

Multi-objective EAs, or MOEAs, are used to solve problems which have multiple, and often conflicting, objectives. A central concept for MOEAs, and multi-objective optimisation in general, is that of a non-dominated solution. This is a solution which is no worse than any of the other solutions within the population when all objectives are taken into account, and the aim of an MOEA is to build and maintain a population of non-dominated solutions that cover all trade-offs between the objectives. This is known as the Pareto optimal front. Exactly how this is achieved varies between MOEAs. However, a well known example is NSGA-II (Non-Dominating Sorting Genetic Algorithm) [31]. Prior to selection, NSGA-II ranks all solutions in terms of dominance: those which are non-dominated are assigned rank 1, those which are only dominated by rank 1 solutions are assigned rank 2, etc. The population is then ordered by rank, and by a measure of crowding distance within ranks, and the first half of the ordered population is copied directly into the next generation. The remainder of the population is then filled by breeding, with parents selected from the higher ranks. Hence, non-dominated solutions are preserved between generations, and new solutions are explored via inter-breeding, resulting in a diverse set of non-dominated solutions that approximate the Pareto optimal front.

The core challenge faced by multi-objective optimization (and absent from single-objective optimization) is: how to rank candidate solutions in a way that leads to effective selection pressure, especially when the entire population (or most of it) may be non-dominated. Another way in which multi-objective optimization differs from single-objective optimization is in the nature of the 'best-so-far solution'. In single-objective optimization the 'best-so-far' solution is trivial to define and to keep track of; in multi-objective optimization the situation is vastly different: the solution is, technically, the entire Pareto front, which is usually a set of solutions, whose cardinality may vary from one to the entire search space. For MOEAs, this leads to certain technical issues which are invariably addressed by maintaining an *archive* of non-dominated solutions; this archive simply keeps track of the 'best-so-far' approximation to the Pareto front, but is also often used as a reservoir for selection of parents. Approaches to the main challenge – how to apply effective selection pressure among the current population – are far more varied. While the approach taken by NSGA-II, as detailed above, is a common and quite successful one, many other styles of MOEA exist, which take different approaches to this central question. In PAES [32], for example, there is only a single 'current' population member. Selection is consequently simplified, however the challenge shifts to the question of whether or not to update the current solution with a newly generated one when the two are non-dominated; PAES makes this decision with the aid of its archive, preferring to explore new areas of the search space than to stay close to solutions already in the archive. Meanwhile, a different breed of MOEAs in this respect is represented by MOEA/D [33]; bypassing the need to distinguish between non-dominated solutions for selection purposes, MOEA/D 'decomposes' a multi-objective problem into many single-objective simplifications of it, each involving a different weighting of the objectives. MOEA/D conducts these single-objective searches in parallel (typically using a local search mechanism), and organises occasional communication between them, as well as book-keeping activities that build and maintain the archive. Effectively, each of MOEA/D's single-objective searches explores a different area of the Pareto front. There are many other approaches, and MOEAs are becoming increasingly used as it becomes recognised that real-world problems are almost invariably multiobjective in nature. Further discussion of the latter point, as well as a first introduction to MOEAs, may be found in [34], while an example of a fairly recent review of MOEAs may be found in [35].

### 3.5 Dynamic Optimisation

So far, our discussion of optimisation has only considered problems in which the search space remains fixed. In many real world problems this is not the case, and various EA approaches are used to handle these situations. Dynamic optimisation is an area in which EAs might be expected to perform relatively well, since the natural diversity present in their populations provides a recovery mechanism that can respond to slow changes in the optimisation landscape. This is especially the case when diversity maintenance techniques are implemented, such as those already discussed in the sections on multimodal and multi-objective optimisation. However, this diversity may be insufficient when the optimisation targets change rapidly or abruptly. A simple solution in this situation is to inject extra diversity into the population when a change is detected, for instance by adapting the variation operators so that larger moves are made. Detection of change can be done by re-evaluating a proportion of the population, looking for significant changes in fitness.

A variety of more elaborate approaches have been developed to handle dynamic optimisation in EAs. An approach inspired by biological systems is to use redundancy in the encoding of a solution. Rather than replacing components of a solution when variation operators are applied, this allows old components to become recessive, i.e. to remain present within the solution but not be expressed during evaluation. Later in the evolutionary process, these components can become reactivated, in effect providing a mechanism to backtrack to previous search points. This is particularly useful when changes in the search space are cyclic. A well-known example is the use of *multiploidy* in GAs [36], where each solution has multiple chromosomes (only one of which is dominant) and variation operators are able to move information between chromosomes. Other approaches to handling dynamic search spaces include predicting change and using multiple populations; see [37] for a recent review.

### 3.6 Coevolving Solutions

Coevolutionary algorithms [38] are motivated by the interactions that occur between species during the course of biological evolution, and the roles these interactions are thought to play in the evolution of complex organisms. Most coevolutionary algorithms use multiple populations, one per *species*. Coevolutionary relationships in biology can be cooperative or competitive. The latter class are particularly well known, and are encapsulated in the idea of predator-prey patterns of evolution, where an arms race between two species can lead to the rapid emergence of complex adaptations. Similar ideas have been explored in EAs, the classic example being the co-evolution of sorting networks and sorting algorithms [39]: the discovery of harder problems (the sorting networks in the first population) leads to selective pressure to discover better solutions (the sorting algorithms in the second population), which leads to selective pressure to discover harder problems, and so on. Competitive coevolution can be used to solve hard problems, and is also useful in circumstances where a fitness function cannot be defined. However, competitive coevolution is known to be difficult to control, and pathological situations can lead to ineffective search. See [38] for a review.

Cooperative coevolution, by comparison, is seen as a useful mechanism for breaking down large problems into more tractable chunks [40]. The idea is that a solution to a problem is divided into sub-components. Each of these sub-components is then evolved in a separate population, with its objective value dependent upon how compatible it is with sub-components being evolved in other populations. Following this, the co-adapted sub-components are then assembled to form a complete solution. In [41], for example, the authors describe how a cooperative coevolutionary variant of DE can be used to solve numerical optimisation problems with up to 1000 variables. Cooperative coevolution can also take place within a single population. An example of this is a Michigan-style learning classifier system (LCS), a form of EA that coevolves a population of rules that can collectively solve difficult problems in classification and machine learning [42].

## 4. Applying Evolutionary Algorithms

### 4.1 Choosing a Methodology

It can be difficult to choose which EA to use for a particular task, since there are many different EAs in common use and relatively little in the way of objective comparative guidance. In practice, it may be necessary to try out different EAs to find out which is the best match to a problem, especially when the problem is poorly understood. However, given whatever is known about the problem at hand, it might be possible to leverage existing understanding of the strengths and weaknesses of particular algorithm frameworks. Some guidance on this matter is available in studies of comparative performance mentioned at the end of Section 4.9.1. It is hoped that Section 4.9.2 also provides useful pointers if the problem is multimodal, multiobjective, dynamic, or unusally large and complex. It is also notable that multi-objective and memetic algorithms, in particular, have become popular for solving difficult real world problems.

## 4.2 Choosing Parameters

EAs invariably have many parameters, and once an algorithm has been selected, it is normal to carry out parameter tuning in order to obtain a better fit between the algorithm and the problem. It can be challenging to obtain optimal parameter settings, since parameters are typically both numerous and not independent of one another. DE, for instance, is notable for having relatively few parameters, and this is often portrayed as a strength of the method. However, EAs are relatively forgiving, and good performance is likely to be possible with non-optimal parameter settings. Nevertheless, guidance is available for choosing the settings of certain parameters [43] and a number of techniques have been developed for automating the choice of parameter settings [44].

## 4.3 Software Tools

| Tool | Language | Summary |
| --- | --- | --- |
| DEAP https://code.google.com/p/deap/ | Python | Distributed Evolutionary Algorithms in Python offers good support for GAs and ESs. Also implements GP and MOEAs. |
| ECJ http://cs.gmu.edu/~eclab/projects/ecj/ | Java | Continuously developed since 1998, Evolutionary Computation in Java has particular strength in GP, but also implements GAs, DE and MOEAs. |
| EO http://eodev.sourceforge.net | C++ | Evolving Objects is an established general purpose EA library with implementations of GAs, GP, ESs and EDAs. |
| EvA2 http://www.ra.cs.uni-tuebingen.de/software/JavaEvA/ | Java | EvA2 is a general purpose EA framework, but has particular strengths in ESs and EDAs, including BOA. |
| HeuristicLab http://dev.heuristiclab.com/ | C# | HeuristicLab implements many of the common EA varieties, and also has support for other population-based and local search metaheuristics. |
| MOEA Framework http://www.moeaframework.org/ | Java | A relatively new EA framework with considerable strength in MOEAs, and multiobjective variants of DE and GP. |
| OpenBEAGLE https://code.google.com/p/beagle/ | C++ | A long established EA framework with good support for GAs and ESs. Also implements GP and NSGA-II. |

**Table 1: Open source EA frameworks**

Tools support is an important issue for many practitioners, and a particular EA methodology is likely to be more appealing if it has a mature supported implementation. Tools support is also important if it is necessary to handcraft a new algorithm to solve a particular problem, and in this situation the language used by the tool may also be a significant concern. Table 1 summarises the features of some of the better known EA tools. GAs are widely supported by all of these. ES support is also widely available, though EvA2 stands out in this regard, with implementations of a wide range of ES variants. DE and EDAs are more recent algorithms, and this is reflected by fewer mature tools. However, EvA2 is again notable for having an implementation of BOA and other EDAs. GP support is offered by a number of these tools, with ECJ implementing a particularly wide range of GP variants. Most also offer support for multimodal and multi-objective approaches, though MOEA Framework stands out for the latter. All of these tools allow custom code to be written. Most use Java or C++, though DEAP is notable as a mature Python implementation and HeuristicLab is available for C# users.

## 5. Case Studies

### 5.1 Evolutionary Algorithms at Large

Now more than half a century since the first appearance of 'EA'-style algorithms in the research literature (widely considered to be [45]), EAs have penetrated almost every area of science and industry, and are regularly used in solving an immense range of optimization problems. To name just a few of the areas in which EAs have had much impact, we can list aeronautical and automotive design [46], bioinformatics and biotechnology [47], chemical engineering [48], creative pursuits [49], finance and investment [50], manufacturing [51], and structural design [52]. To select a small number of case studies could not serve to characterise the true diversity of applied EAs. We therefore duck that challenge, and take the liberty of concluding this chapter by providing two case studies from the authors' recent work, illustrating how some EA approaches are being applied to diverse and challenging optimisation problems in just one corner of science.

### 5.2 Using Niching and Co-Evolution to Understand Gene Regulation

Understanding gene expression is fundamental to understanding living processes. The expression of each gene in an organism is determined by the binding of special proteins, called transcription factors, within a region of DNA upstream of its coding region. In higher organisms, such as humans, this regulatory region typically contains around 5-10 transcription factor binding sites. Characterising these bindings sites, both individually and in combination, is a fundamental part of reconstructing (and ultimately controlling) the genetic networks that underlie biological function.

Identifying binding sites is often reduced to an optimisation problem that involves constructing a matrix model of the occurrence of each DNA base at each position within a short region of DNA. Candidate solutions to this problem can be evaluated by scanning them along the regulatory regions of groups of genes which are known to be expressed at a certain time or within a certain cell type, looking for matches to patterns embedded in the sequences. In most cases, this is a multimodal problem, since multiple binding sites are likely to be relevant to a particular regulatory context, and it is important to be able to identify these different optima. However, these binding sites can vary quite considerably in their degree of conservation, meaning that the objective values of different optima also vary considerably.

Identifying and preserving different modes within a multimodal search space is a challenging problem, especially when they have different relative finesses. In Section 4.9.3, we discussed the idea of niching within the populations of EAs as a means of addressing multimodal problems. In [13] we used an EA furnished with a particular form of niching, termed population clustering, that uses a clustering algorithm to identify and preserve the different modes present within an evolving population of solutions. This allowed a number of different binding sites to be characterised and preserved during a single run. Compared to other forms of niching, it also had the benefit of explicitly identifying these different groups of solutions, allowing the progress of search to be visualised and for clustering parameters to be dynamically modified by an expert user. Even in the absence of dynamic modification, however, this was effective at identifying the clusters of binding sites within comparatively long regions of DNA.

Identifying binding sites is only one part of the problem. Another important aim is to understand the interactions between different binding sites during gene regulation. In [53], we used coevolution to explore solutions to this problem. This involved coevolving two populations, the first containing matrix models of binding sites, the second containing Boolean expressions describing their co-occurrence within binding regions. Members of the binding site population were used as leaves within the Boolean expressions. In essence, the problem of identifying binding sites and their co-expression was decomposed into two problems which were then solved in parallel, using coevolution to provide feedback between the populations. In comparison to a more traditional approach, which would involve sequentially learning binding sites and then learning their interactions, this allows search to be directed towards solutions that interact well with other solutions during the course of search. Such solutions, in turn, are more likely to be meaningful. This approach proved effective for reverse engineering the regulatory rules underlying differential gene expression within tissues. More surprisingly, it also provided a mechanism for solving harder instances of the single binding site optimisation problem, with co-evolution provided a means of implicitly decomposing the matrices associated with these harder problems.

**5.3 Evolving Classifiers for Parkinson's Disease Diagnosis**

This second case study concerns a problem in which the optimal representation for solutions is not clear in advance, requiring experimentation with different kinds of solution representation. As discussed earlier, EAs are entirely flexible in this regard. The problem involves building diagnostic classifiers for Parkinson's disease. These are required to reach their decision based on time series movement data recorded whilst patients and age-matched control were undergoing clinical assessments of motor function. Motor aspects of Parkinson's are incompletely understood, making it unclear what kind of features of the data are important.

To address this, we used EAs to explore two different relatively unconstrained classifier models. Initially we considered a GP-based approach, using it to discover mathematical expressions that describe over-represented patterns of movement embedded in short segments of the time series data, i.e. a form of symbolic regression [54]. An advantage of this approach is that the resulting expressions were relatively interpretable, allowing us to gain insight into the basis of classifications, and then pass this information on to our clinical partners. In particular, analysis of the evolved expressions identified specific aspects of the closing phase of a 'finger tap' movement as highly discriminatory of Parkison's disease patients versus control. These factors alone were indeed more discriminatory than standard metrics, however overall classification performance was not quite as good as that of trained clinicians.

We then considered a more unusual method of representing programs, artificial biochemical networks [55]. These are abstract executable models of the networks of biochemical interactions that underlie the function of biological cells. In a nutshell, they attempt to capture the representation which biological evolution has selected to optimise complex behaviours, with the hypothesis that this makes them particularly suitable for use with EAs. Meanwhile, the use of an EA to search through the space of possible artificial biochemical network classifiers represents, in itself, a major and commonly understood strength of EAs: they can be tailored and deployed effectively with relative ease despite the complexity and diversity of the structure they are being used to evolve. By using this approach, we were now able to find classifiers that produced comparable performance to that of trained clinicians [56]; with accuracy at around 90% overall, accuracy was comparable to the diagnostic accuracies found in clinical diagnosis, and significantly higher than those found in primary and non-expert secondary care.